\documentclass[preprint,12pt,3p]{elsarticle}

\usepackage{amssymb}
\usepackage{xfrac}

\usepackage[export]{adjustbox}
\usepackage{circuitikz}
\usetikzlibrary{shapes,arrows,babel}
\usepackage{tikzscale}
\usepackage{listings}
\usepackage{listingsutf8}
\usepackage{amsmath}
\usepackage{ragged2e}
\usepackage{subcaption}
\usepackage{steinmetz}
\usepackage{float}

\usepackage{array,multirow,graphicx}
\usepackage{amsmath,amssymb,bm,bbm}
\DeclareMathOperator*{\argmax}{arg\,max}
\usepackage{hyperref}
\usepackage{blindtext}
\usepackage{algorithm,algpseudocode}
\algnewcommand\algorithmicforeach{\textbf{for each}}
\algdef{S}[FOR]{ForEach}[1]{\algorithmicforeach\ #1\ \algorithmicdo}
\algdef{S}[FOR]{ForEachP}[1]{\algorithmicforeach\ #1\ \textbf{do in parallel}}
\usepackage[usestackEOL]{stackengine}

\journal{Emergent cooperation through mutual information maximization - arXiv preprint}

\begin{document}

 
 

\begin{frontmatter}

\title{Emergent cooperation through mutual information maximization}

\author{Santiago Cuervo and Marco Alzate\\ \{\texttt{dscuervog@correo.udistrital.edu.co, malzate@udistrital.edu.co}\} \\ \footnotesize \textit{Universidad Distrital Francisco Jos\'e de Caldas} \normalsize}

\begin{abstract}
With artificial intelligence systems becoming ubiquitous in our society, its designers will soon have to start to consider its social dimension, as many of these systems will have to interact among them to work efficiently. With this in mind, we propose a decentralized deep reinforcement learning algorithm for the design of cooperative multi-agent systems.  The algorithm is based on the hypothesis that highly correlated actions are a feature of cooperative systems, and hence, we propose the insertion of an auxiliary objective of maximization of the mutual information between the actions of agents in the learning problem. Our system is applied to a social dilemma, a problem whose optimal solution requires that agents cooperate to maximize a macroscopic performance function despite the divergent individual objectives of each agent. By comparing the performance of the proposed system to a system without the auxiliary objective, we conclude that the maximization of mutual information among agents promotes the emergence of cooperation in social dilemmas.
\end{abstract}


\end{frontmatter}

\section{Introduction}
\label{intro}

Artificial intelligence (AI) systems are nowadays ubiquitous in our society, as several AI-based technologies have gone mainstream and are now an essential part of the workings of our phones, social media, search engines, online stores, streaming services, and many other aspects of our day to day lives. This trend is likely to continue, and to become even more pervasive with the advent of technologies like self-driving cars, that will put AI systems straight into our physical reality \cite{AI100}.


As more and more of these artificial agents populate our world, we will soon have to start to consider its social dimension, since they will face \textit{social dilemmas} similar to the ones we humans encounter, and which, if not properly handled, would act in detriment of their benefit to us. For instance, a set of self-driving cars selfishly trying to cross an intersection as fast as possible to minimize their traveling times, regardless of others, would result in a \textit{prisoner's dilemma}-like problem in which traffic congestion and probability of accidents increases \cite{Schwarting19a}. In this scenario, we  would like instead that our agents coordinate with each other to improve the traveling times of the system as a whole. Such problems, where multiple agents, with possibly conflicting individual objectives, seek to jointly maximize a macroscopic performance function, are termed \textit{Cooperative Multi-Agent Systems} (CMAS) \cite{Panait05a} and are the focus of this work.

In this paper we propose an algorithm for the design of CMAS using \textit{deep reinforcement learning} (DRL), a combination of \textit{reinforcement learning}, an area of machine learning where an agent learns by interacting with a dynamic environment \cite{Sutton98a}, and \textit{deep learning}, a set of techniques based on neural networks which excels at dealing with high dimensional raw data, such as images and speech, and that is responsible for most of the recent milestones achieved in AI research \cite{Lecun15a}. The application of DRL to CMAS has been attracting increasing research interest in recent years, but although many algorithms have been proposed \cite{Hernandez19a}, most of them resort to centralized learning to achieve cooperation, an strategy that is not feasible in many practical problems of an inherently distributed nature \cite{Panait05a}.

To tackle the problem of decentralized learning, we design the individual learning process of the agents such that cooperation is an emergent property in the system, rather than a hard wired feature. We argue that correlation between the actions of agents is a key ingredient of cooperative systems, as it would measure coordination, and based on this, we propose a DRL algorithm that seeks to maximize a differentiable estimate of the \textit{mutual information} (MI), a nonlinear correlation index, between the actions of the agents. We hypothesize that by promoting the maximizing of MI as part of the learning problem of each agent, coordination, and possibly cooperation, could emerge in the system.

The maximization of MI in agent-centric problems has been previously treated in the literature on \textit{empowerment} \cite{Klyubin05a}, where the MI between the agent and the environment is proposed as an universal measure of control. Empowerment has been applied in single-agent DRL algorithms, for instance, in \cite{Mohamed15a} and \cite{Kumar18a}, estimates of the MI are used as an intrinsic reward to perform empowerment-based reasoning. Our work is also closely related to the one in \cite{Jaques18a}, where an estimate of the point-wise MI between the actions of agents is proposed as an intrinsic reward to model social influence and foster cooperation, but their approach is poorly scalable to large systems, since its estimation of the MI requires a model of the whole population. We prescind of the need of such model by considering just the actions of other agents in the vicinity of the learner, and encoding them as a continuous variable whose dimension does not depend on the number of agents. Therefore, our algorithm allows for large populations and even populations whose size changes in time. Also, because our MI estimator is differentiable and we optimize it directly using a gradient ascent algorithm, it is reasonable to think that, with a good quality estimator, this approach would provide a better learning signal than if using an intrinsic reward.


This paper is organized as follows. We start by proposing a quantitative definition of cooperation based on the correlation between the actions of agents in section \ref{quant_coop}. Next, in section \ref{problem_setup}, we define the learning problem of an agent that intuitively could maximize such quantity, and in section \ref{agent_design} the design of a DRL agent to approximately solve it. In section \ref{cg} we describe the \textit{commons game}, a social dilemma of renewable resource consumption, to which we apply our algorithm according to the experimental setup detailed in section \ref{experiments}. The obtained results are shown in section \ref{results}, and its implications discussed in section \ref{discussion}. Finally, we present our conclusions in section \ref{conclusions}.

\section{An index of cooperation in multi-agent systems}
\label{quant_coop}
Several definitions of cooperation have been proposed in the literature from the perspective of diverse scientific fields, such as evolutionary biology \cite{Lindenfors17a}, game theory \cite{Murphey02a} and information theory \cite{Griffith14a}. These definitions share several ideas, such as the macroscopic nature of cooperation, being a feature of a set of entities rather than of individuals, the existence of a common objective across the set of entities, and the idea that is the relationships among the elements of the set that results in an improvement towards the objective. Here, in line with these ideas, we define cooperation in the context of multi-agent systems as an 

\begin{center}
\emph{attribute of a coordinated set of actions in a multi-agent system that causes the improvement of the system performance}    
\end{center}, and define a very simple scalar index to quantify it. The coordination of the actions means that these are not independent, and can be quantified with a correlation index. Let $J$ be a system-level performance index, and let $\rho$ be a non-negative positive correlation index between the actions of the agents, then we define a \textit{cooperation index}, $\psi$, as

\begin{equation}
    \psi = \rho \cdot J
    \label{eq:coop_idx}
\end{equation}

The index $\psi$ will be high for highly correlated actions that result in a high performance, and it will be zero for independent actions even if they result in high performance.

\section{Problem setup}
\label{problem_setup}
The interaction of an agent with its environment in reinforcement learning is formalized as a \textit{Markov Decision Process} (MDP) \cite{Sutton98a}. An MDP is defined by the tuple $(\mathcal{S}, \mathcal{A}, R, T, \gamma)$. Where $\mathcal{S}$ is the set of all possible states. $\mathcal{A}$ is the set of all possible actions. The transition function $T: \mathcal{S} \times \mathcal{A} \times \mathcal{S} \to [0, 1]$ defines the probability of a transition from the state $\mathbf{s} \in \mathcal{S}$ to the state $\mathbf{s}' \in \mathcal{S}$ given an action $\mathbf{a} \in \mathcal{A}$. The reward function $R: \mathcal{S} \times \mathcal{A} \times \mathcal{S} \to \mathbb{R}$ defines the immediate reward $r \in \mathbb{R}$ that an agent would receive given that executes action $\mathbf{a}$ in state $\mathbf{s}$ and is transitioned to state $\mathbf{s}'$. Finally, $\gamma \in [0, 1]$ is the discount factor that balances the trade-off between short-term and long-term rewards.

Solving an MDP consists in finding a mapping from states to actions, termed policy, $\pi: \mathcal{S} \to \mathcal{A}$, where the optimal policy, $\pi^*$, is defined as:
\begin{equation}
    \pi^* = \argmax_\pi V(\mathbf{s}_0, \pi), \; \forall \mathbf{s_0} \in \mathcal{S}
\end{equation}, with $V$ being the state value function, defined as the expected long-term payoff of being in an initial state, $\mathbf{s}_0$, if actions are chosen according to the policy, $\pi$:

\begin{equation}
    V(\mathbf{s}_0, \pi) = \mathbb{E}_{\mathbf{s}_{t+1} \sim T}\left[\sum_{t=0}^{\infty} \gamma^t R(\mathbf{s}_t, \pi(\mathbf{s}_t), \mathbf{s}_{t+1}) \right]
\end{equation}

In a multi-agent system the MDP turns into a \textit{Markov Game} (MG), where the transition and reward functions depend on the joint action of all the agents \cite{Hernandez19a}. That is, $\mathcal{A}$ is redefined as the set of all possible joint actions, $\mathcal{A} = \mathcal{A}^{(1)} \times ... \times \mathcal{A}^{(n)}$, with $\mathcal{A}^{(i)}$ being the set of all possible individual actions of the $i$-th agent, for $i=1,...,n$, in a system of $n$ agents. From a single agent perspective this renders the process non-stationary, since the value function, and thus the optimal policy, depend on the policies of all the other agents in the system that are also changing in time as they learn. Most of the approaches proposed in the literature to deal with non-stationarity in systems of multiple learners resort to centralized strategies, where global information is used by a single learner to learn value and/or policy functions for the whole system \cite{Hernandez19a}.

Here, our focus is on problems that must be solved on a distributed manner, and therefore, centralized learning is not feasible. We consider each agent as an independent learner that at each time step receives a local observation, $\mathbf{o} \in \mathcal{O}$, where $\mathcal{O}$ is the set of all possible observations, and a local reward, $r$. Using only this local information it has to learn a policy that maximizes the global long-term payoff of the system by coordinating with others.

Inspired by the definition of cooperation given in section \ref{quant_coop}, we hypothesize that by promoting the maximization of correlation between the actions of agents, along with the maximization of individual value functions, we can guide the learning process towards the desired regions of the search space that define highly rewarding and highly correlated policies, and that, following the definition, would likely result in cooperative behaviors and improved global performance. 

Let the policy $\pi_{\theta}$ be a neural network parameterized by $\theta$, then the learning problem of the $i$-th agent is formulated as

\begin{equation}
    \max_\theta \left[\, c_1 \, V(\mathbf{s}_0, \pi_\theta) + c_2 \, \mathbb{E}_{(\mathbf{a}, \mathbf{a}^{(-i)}) \sim \pi_\theta \pi^{(-i)}}\, \rho(\mathbf{a}, \mathbf{a}^{(-i)}) \right], \; \forall \mathbf{s}_0 \in \mathcal{S}
    \label{eq:ag_prob}
\end{equation}, where $\rho$ is a non-negative correlation index between the actions of the agent, $\mathbf{a}$, and the joint action of other agents in the system, $\mathbf{a}^{(-i)}$, determined by its joint policy, $\pi^{(-i)}$.

\section{Agent design}
\label{agent_design}
\subsection{Agent architecture}
\begin{figure*}
    \centering
    \includegraphics[width=0.75\textwidth]{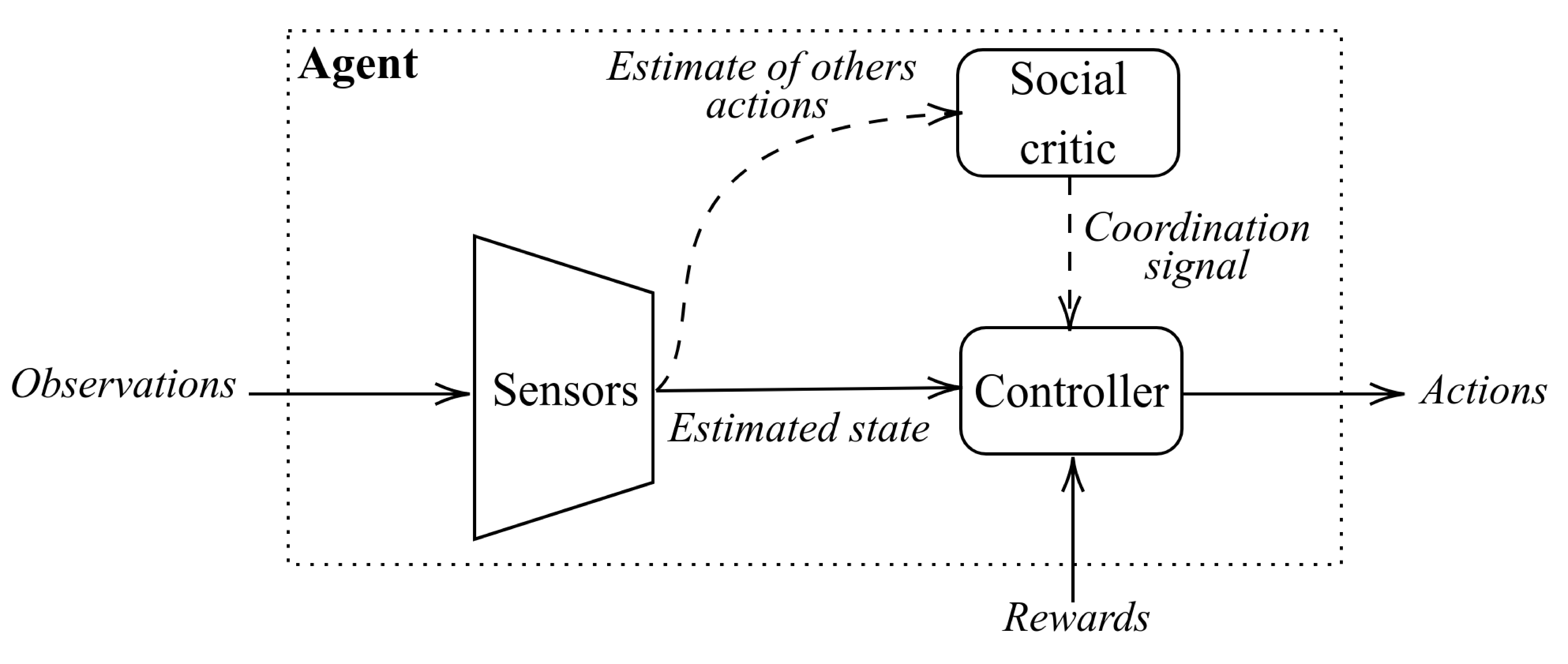}
\caption{Agent architecture. The sensors (trained offline) extract relevant information about the environment and about other agents that it is used by the social critic and the controller (trained online) to guide the agent towards coordinated and highly rewarding policies.}
\label{fig:agent_arch}       
\end{figure*}

The overall architecture of the agent designed to approximately solve the optimization problem described in equation \ref{eq:ag_prob} is illustrated in figure \ref{fig:agent_arch} and it is inspired by the modular design proposed in \cite{Ha18a}, where the agent is composed of a feature extraction component that is trained offline, and a decision making component that is trained online, as the agent interacts with the environment. 

The agent is composed of three functional modules. The \textit{sensors} receive the observations from the environment, that are assumed to be high dimensional and unstructured, and extract relevant information from it to produce estimates of the state of the environment and of the actions of other nearby agents. The \textit{social critic} receives the estimate of others actions and estimate its MI with the actions of the agent. Just like in \textit{actor-critic} algorithms the critic component guides the learning dynamics of the policy towards high rewards \cite{Konda03a}, the gradient of the mutual information estimated by the social critic is used during learning to guide the agent towards more coordinated behavior with its peers. Finally, the \textit{controller} implements the policy of the agent. It receives as input the states estimated by the sensors, produces as output actions, and updates the policy using the observed rewards and the signal produced by the social critic. Each module is composed of multiple neural networks that are trained using a pipeline of several stages of machine learning. Below we described them in detail.

\subsubsection{Sensors}
At each time step the agent receives a high dimensional observation, $\mathbf{o}$, typically, a 2D image that is part of a video sequence. We use a neural network, $E_x$, to learn a compressed representation of each observed input frame. $E_x$ is implemented as the encoder component of an \textit{undercomplete autoencoder} \cite[pp. 500-501]{Goodfellow16a}, that receives $\mathbf{o}$ as input and produces as output a code $\mathbf{x}$. The training process consists then on minimizing the loss function,

\begin{equation}
    L_{E_x}(\mathbf{o}, D_x(\mathbf{x}))
    \label{eq:e1_loss}
\end{equation} ,where $L_{E_x}$ penalizes $D_x(\mathbf{x})$ for being dissimilar from $\mathbf{o}$, and $D_x$ is the decoder component of the autoencoder. 

A second encoder, $E_y$, is used to extract the information related to other agents from the code $\mathbf{x}$ to another code, $\mathbf{y}$. This information is later used by the social critic to estimate the actions of other agents. $E_y$ is also the encoder component of an undercomplete autoencoder, trained to minimize the loss function,

\begin{equation}
    L_{E_y}(\mathbf{o}^{(-i)}, D_y(\mathbf{y}))
    \label{eq:e2_loss}
\end{equation} ,where $\mathbf{o}^{(-i)}$ is the input containing only the information related to other agents in the original observation. $L_{E_y}$ penalizes $D_y(\mathbf{x})$ for being dissimilar from $\mathbf{o}^{(-i)}$, and $D_y$ is the decoder component of the autoencoder.

While $E_x$ and $E_y$ compress what the agent sees at each time step, we also want to compress what it sees over time. This is necessary since the agent must deal with a partially observable system, and hence, requires memory to estimate the state of the environment and be able to take optimal decisions \cite{Aberdeen03a(revised)}. We use a \textit{recurrent neural network} (RNN) \cite[pp. 367-415]{Goodfellow16a}, $M$, that serves as a memory for the agent by storing in its state information about past observations. At each time step, it receives as input the current compressed observation, $\mathbf{x}_t$ and its current state, $\mathbf{h}_{t}$, and outputs its new state, $\mathbf{h}_{t+1}$. The \textit{estimated state} of the system, $\hat{\mathbf{s}}_t$, is then defined as the concatenation of the present compressed observation and memory state: 

\begin{equation}
    \hat{\mathbf{s}}_t = [\mathbf{x}_t, \mathbf{h}_t].
\end{equation}

With the intention of reducing the complexity of the learning problem, and considering computational costs, we use as memory an \textit{Echo state network} \cite{Jaeger01a}, a kind of RNN whose weights are fixed after initialization.

\subsubsection{Social critic}
The social critic uses the estimator proposed in \cite{hjelm18a} to approximate the MI between the action of the agent in the current time step, $\mathbf{a}_t$, and the joint action of other agents in its vicinity, $\mathbf{a}^{(-i)}_{t}$. However, it does not do so directly, because, with practical considerations in mind, we use proxies for both variables.

To be able to maximize the MI between agents with a gradient-based optimizer, we want our representation of the MI to be differentiable with respect to the parameters of the policy. In this work we consider a discrete action set, therefore, we assume an stochastic policy that defines a probability mass function over the action space conditional to the state, $\pi_\theta: \mathcal{S} \to [0, 1]^{|\mathcal{A}|}$, and estimate the MI between the vector of probabilities, $\mathbf{p}$, and the actions of other agents. In problems with a continuous action set, the actions could be directly used with the estimator.

We would also like to make our approach scalable to populations of any size and variable in time, but we are limited by the dimension of $\mathbf{a}^{(-i)}$. As the size of the population grows it also does the complexity of the learning problem for the estimator because it deals with a higher dimensional input. The dimension of the input also should not change in time, as it would happen with a variable population size. To work around this, we make the assumption that the actions of others agents in the current time step can be approximately inferred from the change between the current and the next time step of the code $\mathbf{y}$, given that it encodes an approximate state of the nearby agents, and use then $\mathbf{y}_{t+1}$ as a proxy to $\mathbf{a}^{(-i)}_{t}$.

The estimator of the MI between $\mathbf{p}_t$ and $\mathbf{y}_{t+1}$ is then defined as,

\begin{equation}
    \begin{split}
    I(\mathbf{p}_{t}, \mathbf{y}_{t+1}) = \max_\omega \left\{- \mathbb{E}_{(\mathbf{p}_{t}, \mathbf{y}_{t+1}) \sim P_{\mathbf{p}_{t}\mathbf{y}_{t+1}}}[-\zeta(-F_\omega(\mathbf{p}_{t}, \mathbf{y}_{t+1}))] \right. \\ \left. - \mathbb{E}_{(\mathbf{p}_{t}, \mathbf{y}_{t+1}) \sim P_{\mathbf{p}_{t}}P_{\mathbf{y}_{t+1}}}[\zeta(F_\omega(\mathbf{p}_{t}, \mathbf{y}_{t+1}))]\right\} 
    \end{split}
\label{eq:mine_loss}
\end{equation}, where $I(\mathbf{p}_{t}, \mathbf{y}_{t+1})$ is the MI between $\mathbf{p}_{t}$ and $\mathbf{y}_{t+1}$, $P_{\mathbf{p}_{t}\mathbf{y}_{t+1}}$ denotes the joint distribution, $P_{\mathbf{p}_{t}}$ and $P_{\mathbf{y}_{t+1}}$ are the marginal distributions, $\zeta$ is the softplus function, and $F_\omega$ is a neural network parameterized by $\omega$.


The expectations in equation \ref{eq:mine_loss} in practice are estimated as averages over samples of the distributions. The samples of the joint distribution are observed by the agent during its interaction with the environment, and $P_{\mathbf{p}_{t}}$ is simply the policy of the agent, but $P_{\mathbf{y}_{t+1}}$ is unknown and needs to be estimated. To do so, we use a neural network, $Y$, to predict $\mathbf{y}_{t+1}$ given the action of the agent, $\mathbf{a}_t$, and the estimated state of the system, $\hat{\mathbf{s}}_t$. $Y$ is trained to minimize the loss function,

\begin{equation}
    L_Y(\mathbf{y}_{t+1},Y(\hat{\mathbf{s}}_t, \mathbf{a}_t))
    \label{eq:s_loss}
\end{equation}, where $L_Y$ penalizes $Y(\hat{\mathbf{s}}_t, \mathbf{a}_t)$ for being dissimilar from $\mathbf{y}_{t+1}$. The samples from $P_{\mathbf{y}_{t+1}}$ are then estimated by averaging out $\mathbf{a}_t$ from $Y$:

\begin{equation}
    \frac{1}{|\mathcal{A}_i|}\sum_{\mathbf{a}_t \in \mathcal{A}_i} Y(\hat{\mathbf{s}}_t, \mathbf{a}_t)
\end{equation}

\subsubsection{Controller}
The controller implements the policy of the agent, $\pi_\theta$, and an estimate of the value function, $\hat{V}$. The policy is trained using the \textit{Proximal Policy Optimization} (PPO) algorithm \cite{Schulman17a}, where the loss function of the policy is defined as:

\begin{equation}
    \begin{split}
        L_\pi = \frac{1}{|\mathcal{T}|}\sum_{(\hat{\mathbf{s}_t}, \mathbf{a}_t, r_{t+1}) \in \mathcal{T}} \left[\min\left(\frac{\pi_\theta(\mathbf{a}_t|\hat{\mathbf{s}_t})}{\pi_{\theta_{old}}(\mathbf{a}_t|\hat{\mathbf{s}_t})}\hat{A}(\hat{\mathbf{s}_t},\mathbf{a}_t),\right.\right.\\\left.\left. clip\left(\frac{\pi_\theta(\mathbf{a}_t|\hat{\mathbf{s}_t})}{\pi_{\theta_{old}}(\mathbf{a}_t|\hat{\mathbf{s}_t})}, 1 - \epsilon, 1 + \epsilon\right) \hat{A}(\hat{\mathbf{s}_t},\mathbf{a}_t)\right)\right]
    \end{split}
    \label{eq:ppo_pi_loss}
\end{equation}, with $\mathcal{T}$ being a set of state-action-reward tuples observed by the agent while interacting with the environment, $\pi_{\theta_{old}}$ is the policy before an update of the policy parameters, $\hat{A}$ is an estimate of the advantage function calculated over a trajectory of length $l$ as

\begin{equation}
    \hat{A}(\hat{\mathbf{s}}_t,\mathbf{a}_t) = r_{t+1} + \gamma r_{t+2} + ...+ \gamma^{l - 1} r_{t + l} + \gamma^{l} \, \hat{V}(\hat{\mathbf{s}}_{t+l}) -\hat{V}(\hat{\mathbf{s}}_t)
    \label{eq:app_adv}
\end{equation}, $clip$ is the clipping function

\begin{equation} 
clip(x, a, b) = 
     \begin{cases}
       a, &\quad x < a,\\
       x, &\quad a \leq x \leq b \\
       b, &\quad x > b \\
     \end{cases}
\end{equation}, and $\epsilon \in (0, 1)$, is a parameter that controls the size of the updates to the policy network.

We use a neural network architecture that shares parameters between the policy and the estimate of the value function, so the loss function to be minimized combines both objectives, and an additional entropy maximization term to encourage exploration as suggested in \cite{Schulman17a}:

\begin{equation}
    L_{PPO} = -c_\pi \, L_\pi + c_V \,L_V(\hat{V}) - c_H \, H(\pi_\theta)
\end{equation}, where,

\begin{equation}
    L_V(\hat{V}) = \frac{1}{|\mathcal{T}|}\sum_{(\hat{\mathbf{s}_t}, \mathbf{a}_t, r_{t+1}) \in \mathcal{T}} \left(\hat{V}(\hat{\mathbf{s}}_t) - \sum_{k = t}^{t + l - 1} \gamma^{(k - t)} r_{k+1}\right)^2
\end{equation}, $H(\pi_\theta)$ is the entropy of the policy, and $c_\pi$, $c_V$ and $c_H$ are constant coefficients. 

Finally, we include in the objective function the MI estimated by the social critic to encourage coordination with other agents, so the learning problem of the controller is formulated as: 

\begin{equation}
    \min_{\theta} \left[L_{PPO} - c_I I(\pi_\theta(\hat{\mathbf{s}}_t), \mathbf{y}_{t+1})\right], \; \forall \hat{\mathbf{s}_t} \in \mathcal{T}, \; \forall \mathbf{y}_{t+1} \in \mathcal{T}_y
    \label{eq:c_loss}
\end{equation}, where $\mathcal{T}_y$ is the set of encoded observations regarding other agents corresponding to each observed estimated state, $\hat{\mathbf{s}_t} \in \mathcal{T}$. 

\subsection{Training algorithm}
\label{training_algo}
The pseudocode of the training algorithm for the multi-agent system is described in algorithm \ref{algo}. First, the weights of all the neural networks that compose each agent are randomly initialized. The memory, $M$, should be initialized such that the spectral radius of its hidden to hidden weight matrix is less than unity \cite{Jaeger01a}. Next, a dataset of observations is obtained by uniformly sampling the set of possible observations, $\mathcal{O}$. This dataset is used to train the encoders $E_x$ and $E_y$ by applying a gradient descent algorithm to minimize equations \ref{eq:e1_loss} and \ref{eq:e2_loss}, respectively. Once the sensors are trained, the interaction of the agents with the environment begins. 

For a maximum of $t_{max}$ time steps (that could be infinity), each agent follows an iterative training procedure. It begins by acquiring experiences by interacting with the environment and with other agents. For a finite number of time steps, $l$, each agent receives an observation, $\mathbf{o}_t$, executes an  action, $\mathbf{a}_t$, according to its current policy, $\pi_\theta$, and receives a reward, $r_{t+1}$. The observed sequence of observations is encoded by the sensors to produce a sequence of estimated states. The sequence of estimated states, actions and rewards, $\{(\hat{\mathbf{s}}_0, \mathbf{a}_0, r_1), ..., (\hat{\mathbf{s}}_{l-1}, \mathbf{a}_{l-1}, r_l) \}$, is then used to train the social critic and the controller. $Y$, $F_\omega$, and $\pi_\theta$ and $\hat{V}$ are trained every $n_Y$, $n_F$ and $n_C$ time steps, respectively, using a gradient descent algorithm. This difference in the frequencies of training was deemed necessary since these functions make use of each of other and it was observed that if they are changing at the same rate, frequently, the learning process would not converge. We suggest using $n_Y \leq n_F < n_C$, such that the policies of the agents change slower than the capacity of the function $Y$ to adapt to it, and be able to give a good estimate to $F_\omega$. Similarly, $\pi_\theta$ should change slower than $F_\omega$, to allow it to converge to a good estimate of the MI to guide the learning process of the policy. 

\begin{algorithm}[h!]
 \caption{}
 \label{algo}
 \begin{algorithmic}[1]
    \State Initialize neural networks parameters
    \State Obtain dataset to train $E_x$ and $E_y$ by uniformly sampling $\mathcal{O}$.
    \State Train $E_x$ by minimizing equation \ref{eq:e1_loss} 
    \State Train $E_y$ by minimizing equation \ref{eq:e2_loss}
    \State $t_{total} = 0$
    \While{$t_{total} \leq t_{max}$}
        \ForEachP{agente}
            \State \parbox[t]{295pt}{Interact with the environment for $l$ time steps to obtain trajectories of observations, actions, and rewards}
            \If{$t_{total} \bmod{n_Y} = 0$}
            \State Train $Y$ to minimize equation \ref{eq:s_loss}
            \EndIf
            \If{$t_{total} \bmod{n_F} = 0$}
            \State Train $F_\omega$ according to equation \ref{eq:mine_loss}
            \EndIf
            \If{$t_{total} \bmod{n_C} = 0$}
            \State Train $\pi_\theta$ and $\hat{V}$ according to equation \ref{eq:c_loss}
            \EndIf
            \State $t_{total} = t_{total} + l$
        \EndFor
    \EndWhile
 \end{algorithmic} 
 \end{algorithm}
 
\section{The commons game}
\label{cg}
\begin{figure*}
    \centering
    \includegraphics[width=0.6\textwidth]{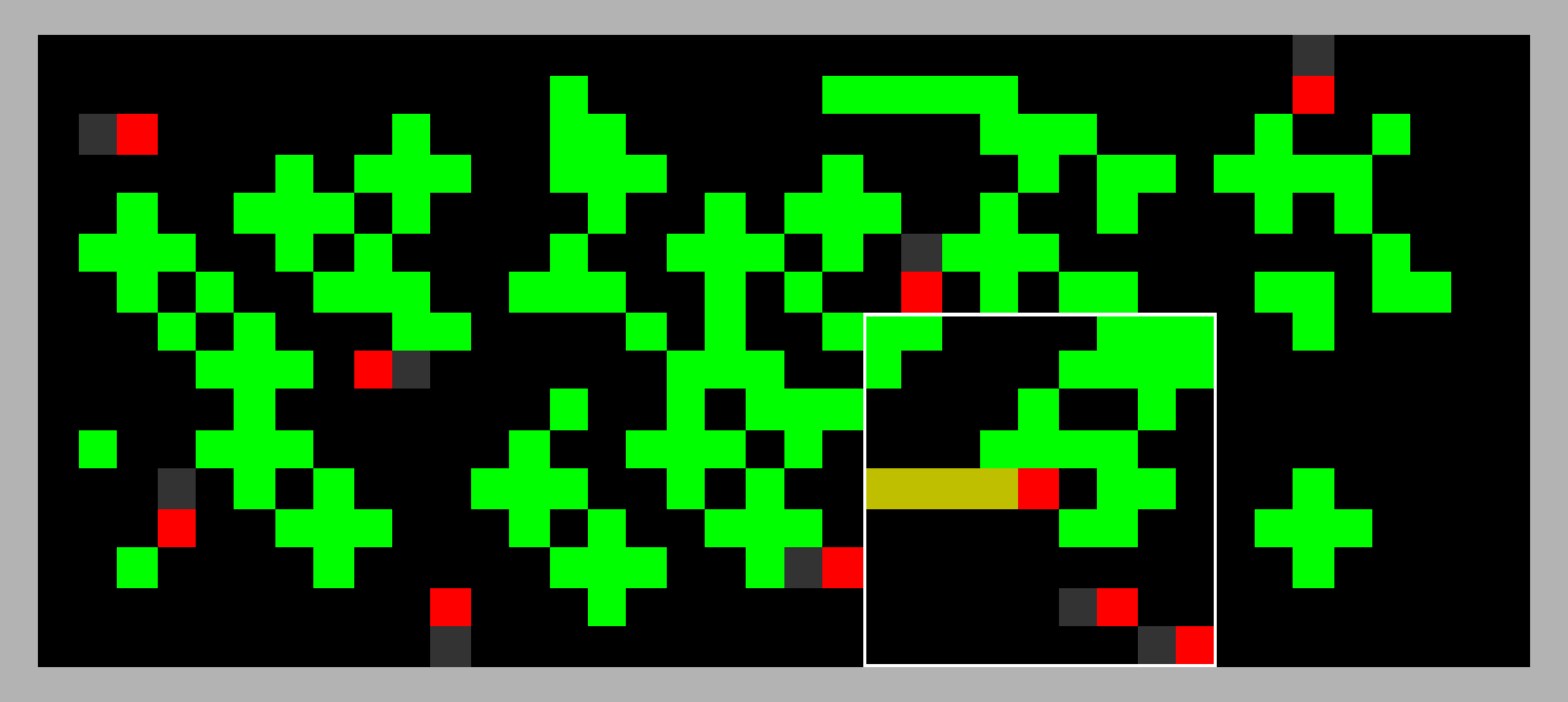}
\caption{A frame of the commons game (own implementation). Agents (red) harvest apples (green). An agent in the southeast of the field shots its beam (yellow) pointing west. Its field of vision is the area contained within the white square.}
\label{fig:harvest}       
\end{figure*}

We applied our algorithm to a \textit{sequential social dilemma} (SSD), a MG with $|S| > 1$ where an agent can get a higher reward by engaging in non-cooperative behavior, but the total payoff per agent is higher if all agents cooperate \cite{Leibo17a}. The chosen SSD is the \textit{commons game} (CG) described in \cite{Perolat17a} and illustrated in figure \ref{fig:harvest}. In the CG a set of agents (red tiles) have to collect apples (green tiles), which are a limited renewable resource. The apple regrowth rate depends on the spatial configuration of the uncollected apples: more nearby apples implies higher regrowth rate. If all apples in a local area are collected then none ever grow back. Agents also can take an offensive action by shooting others with a beam (yellow tiles), which temporally removes them from the game. This reduces the load on the resource by diminishing the effective population size, and enables the aggressive agents to selfishly exploit the resource without depleting it. Cooperation in the CG is achieved when agents coordinate between them to harvest apples in a sustainable way, such that the resource is not depleted, and every agent in the system gets roughly the same amount.

Our CG implementation \footnote{Source code available at: \url{https://github.com/tiagoCuervo/CommonsGame}} uses the map depicted in figure \ref{fig:harvest} with $n=10$ agents and the following features:

\begin{itemize}
    \item Agents have an agent-centered field of vision of radius 4, such that $\mathbf{o} \in \mathbb{R}^{9\times9}$ is an image of the surroundings of the agent. Each agent appears blue in its own field of view, and red in the field of view of other agents.
    \item There are eight possible actions: stay still, go up, go down, go left, go right, turn left, turn right, and shoot beam.
    \item The beam extends within the vision field in the direction the agent is looking and has a width of 1 square. Any agent that is in the path of the beam is removed for 25 time steps.
    \item For every collected apple the agent receives a reward of $r=1$.
    \item At any given time step, a collected apple has a probability $p_r$ of respawning, dependent on the number of apples in a vicinity of radius 2, $n_a$:
\end{itemize}

    \begin{equation}
        p_r(n_a) =
        \begin{cases}
           0, &\quad n_a = 0 \\
           0.01, &\quad 0 < n_a \leq 2  \\
           0.05, &\quad 2 < n_a \leq 4  \\
           0.1, &\quad n_a > 4
         \end{cases}
    \end{equation}
\begin{itemize}
    \item The game finishes if all the apples in the field are harvested.
\end{itemize}



\section{Experiments}
\label{experiments}
\subsection{Experimental setup}
The performance of a system trained with our algorithm, here after called \textit{correlation maximizing system} (CMS), is compared with a baseline system trained with the standard PPO algorithm, this is, by setting $c_I=0$ in equation \ref{eq:c_loss}. 30 independent experiments for both systems are conducted, each consisting of 10 million time steps of interaction with the environment. The architecture of the neural networks composing an agent is described in table \ref{tab:sensors_arch}. These were defined so that most of the model complexity would reside in the sensors and the social critic, mimicking what was proposed in \cite{Ha18a}. No further experimentation was done to look for optimal architectures. All the neural networks are trained using the \textit{Adam} optimizer \cite{Kingma14a} with the parameters presented in table \ref{tab:trn_params}.

The dataset to train $E_x$ was obtained by executing a population of 10 agents with random uniform policies on the environment for $1.28 \times 10^6$ time steps, and storing the frames seen by each agent. This dataset was divided in a training set of $1\times10^7$ samples and a validation set of $2.8 \times 10^6$ samples. The dataset to train $E_y$ was obtained simply by masking the pixels of the own agent, its sight, apples and wall from the dataset of $E_x$, leaving just the pixels corresponding to other agents. The decoders were trained as classifiers for each pixel, hence the Softmax activation in the output layer, where classes correspond to the possible values that a pixel can take in the CG: agent(blue), other agents(red), apple (green), beam(yellow), agent sight (dark gray), and wall (light gray). This allowed the decoders to make perfect reconstructions. The sensors are only trained once, offline, and are later used by all the agents in the system. The dataset for the social critic and controller consist of $l=1000$ time steps of interaction with the environment, as described in algorithm \ref{algo}.

\begin{table}[!t]
\caption{Architecture of the neural networks that compose the agent. The numbers in the layer column indicate the order from input to output. Two layers with the same number process the same input.}
\label{tab:sensors_arch}       
\centering
\begin{tabular*}{\textwidth}{@{}c@{\extracolsep{\fill}}lcc@{}}
\hline\noalign{\smallskip}
 Component & Layer & Dimension & Activation function \\ \hline\noalign{\smallskip}
 \multirow{6}{*}{\vspace{0.5cm} ${E}_x$} & 1. Input & $9 \times 9 \times 7$ & $\cdot$ \\
 & 2. Convolution & $(3 \times 3, 16)$ & ReLU \\
 & 3. Reshape & 784 & $\cdot$ \\
 & 4. Fully connected & 32 & ReLU \\
 & 5. Fully connected & 32 & Linear \\ \hline\noalign{\smallskip}
 \multirow{6}{*}{\vspace{0.5cm} ${D}_x$} & 1. Fully connected & 32 & ReLU \\
 & 2. Fully connected & 784 & ReLU \\
 & 3. Reshape & $7 \times 7 \times 16$ & $\cdot$ \\
 & 4. Transposed convolution & $(3 \times 3, 7)$ & SoftMax \\ \hline\noalign{\smallskip}
 \multirow{6}{*}{\vspace{1cm} ${E}_y$} & 1. Input & 32 & $\cdot$ \\
 & 2. Fully connected & 512 & ReLU \\
 & 3. Fully connected & 128 & ReLU \\
 & 4. Fully connected & 16 & Linear \\ \hline\noalign{\smallskip}
 \multirow{6}{*}{\vspace{1cm} ${D}_y$} & 1. Fully connected & 128 & ReLU \\
 & 2. Fully connected & 784 & ReLU \\
 & 3. Reshape & $7 \times 7 \times 16$ & $\cdot$ \\
 & 4. Transposed convolution & $(3 \times 3, 5)$ & SoftMax \\ \hline\noalign{\smallskip}
 \multirow{2}{*}{\shortstack{$M$}} & 1. Input & 32 & $\cdot$ \\
 & 2. Recurrent & 512 & $\tanh$ \\ \hline\noalign{\smallskip}
 \multirow{4}{*}{\shortstack{$Y$}} & 1. Input & 544 & $\cdot$ \\
 & 2. Fully connected & 128 & ReLU \\
 & 3. Fully connected & 128 & ReLU \\
 & 4. Fully connected & 16 & Linear \\ \hline\noalign{\smallskip}
 \multirow{4}{*}{\shortstack{$F_\omega$}} & 1. Input & 16 & $\cdot$ \\
 & 2. Fully connected & 32 & ReLU \\
 & 3. Fully connected & 32 & ReLU \\
 & 4. Fully connected & 1 & Linear \\
 \hline\noalign{\smallskip}
 & 1. Input & 544 & $\cdot$ \\
 $\pi_\theta$ & 2. Fully connected & 8 & SoftMax \\
 $\hat{V}$ & 2. Fully connected & 1 & Linear \\
 \hline\noalign{\smallskip}
\end{tabular*}
\end{table}

\begin{table}[!t]
\caption{Parameters of the training algorithm for each component.}
\label{tab:trn_params}       
\centering
\begin{tabular*}{\textwidth}{@{}c@{\extracolsep{\fill}}lc@{}}
\hline\noalign{\smallskip}
Component & Parameter & Value \\ \hline\noalign{\smallskip}
\multirow{5}{*}{\shortstack{$E_x$ and $E_y$}}
& Training set size & $1\times10^7$ \\
& Batch size & 128 \\
& Number of epochs & 10 \\
& Learning rate & $1 \times 10^{-3}$ \\
\hline\noalign{\smallskip}
\multirow{4}{*}{\shortstack{$Y$}}
& Training set size & 1000 \\
& Batch size & 32 \\
& Number of epochs & 10 \\
& Learning rate & $5 \times 10^{-4}$ \\
\hline\noalign{\smallskip}
\multirow{4}{*}{\shortstack{$F_\omega$}}
& Training set size & 1000 \\
& Batch size & 32 \\
& Number of epochs & 10 \\
& Learning rate & 0.01 \\
\hline\noalign{\smallskip}
\multirow{7}{*}{\shortstack{$\pi_\theta$ and $\hat{V}$}}
& Training set size & 1000 \\
& Batch size & 32 \\
& Number of epochs & 5 \\
& Learning rate & $1 \times 10^{-4}$ \\
& $c_V$ & 0.5 \\
& $c_I$ & 0.1 (0.0 for the baseline system) \\
& $c_H$ & 0.01 \\
\hline\noalign{\smallskip}
\end{tabular*}
\end{table}

\subsection{Performance indices}
To evaluate each system we use the macroscopic indices for the CG proposed in \cite{Perolat17a}, designed to characterize the strategies of the whole population of agents. We also define the cooperation index proposed in section \ref{quant_coop} for the specific case of the CG. Let $G^{(i)}$ be the total payoff obtained by agent $i$ over a trajectory of $l$ time steps,

\begin{equation}
    G^{(i)} = \sum_{t=0}^l r^{(i)}_{t+1}
\end{equation}, and let $\mathbbm{1}$ be the indicator function,

\begin{equation}
    \mathbbm{1}(x) = \begin{cases}
       1, &\quad \text{if $x$ is true}\\
       0, &\quad \text{otherwise}\\
     \end{cases}
\end{equation}, then the five performance indices are defined as follows.

\subsubsection{Utilities}

Is the average over agents of the obtained payoff:

\begin{equation}
    U = \frac{1}{n} \sum_{i=1}^n G^{(i)}
\end{equation}

\subsubsection{Equity}

Measures the dispersion of the distribution of payoff within agents:
\begin{equation}
E = 1 - \frac{\sum_{i=1}^{n} \sum_{j=1}^{n} |G^{(i)} - G^{(j)}|}{2n \sum_{i=1}^n G^{(i)}}
\end{equation}

\subsubsection{Peace}
Is the average time without out of game agents:

\begin{equation}
    P = 1 - \frac{\sum_{i=1}^{n} \sum_{t=1}^{l}\mathbbm{1}(\mathbf{o}^{(i)}_{t} = \mathbf{o}_{to})}{nl}
\end{equation}, where $\mathbf{o}^{(i)}_{t}$ is the observation of the $i$-th agent, and $\mathbf{o}_{to}$ is the observation that an agent receives when is impacted by the time out beam.

\subsubsection{Sustainability}

Is the cumulative sum of apples during the trajectory:

\begin{equation}
    S = \sum_{t=1}^l \sum_{i} \mathbbm{1}(\text{$\mathbf{s}^{(i)}_{t} = $ apple})
\end{equation}, where $\mathbf{s}^{(i)}_{t}$ is the $i$-th pixel of the state in the time step $t$.

\subsubsection{Cooperation}
The cooperation index in equation \ref{eq:coop_idx} is defined for the CG as:

\begin{equation}
    \psi = \frac{\bar{I}}{\bar{H}} \, U
\end{equation}, where the system-level performance index, $J$, is the utility, $U$, and the correlation index, $\rho$, is the ratio between the estimated average mutual information (EAMI),
\begin{equation}
    \bar{I} = \frac{1}{n} \sum_{i=1}^n I_i
\end{equation}, and the average entropy of the policies in the system,
\begin{equation}
    \bar{H} = \frac{1}{n} \sum_{i=1}^n H(\pi^{(i)}_{\theta})
\end{equation}, with $\pi^{(i)}_{\theta}$ being the policy of the $i$-th agent, and $I$ being the estimated MI (equation \ref{eq:mine_loss}) for the trajectory of the $i$-th agent. The normalization by the average entropy is made in order to eliminate the dependence of the correlation index from the uncertainty in the system.

\section{Results}
\label{results}
\begin{figure*}
    \centering
    \includegraphics[width=0.495\textwidth]{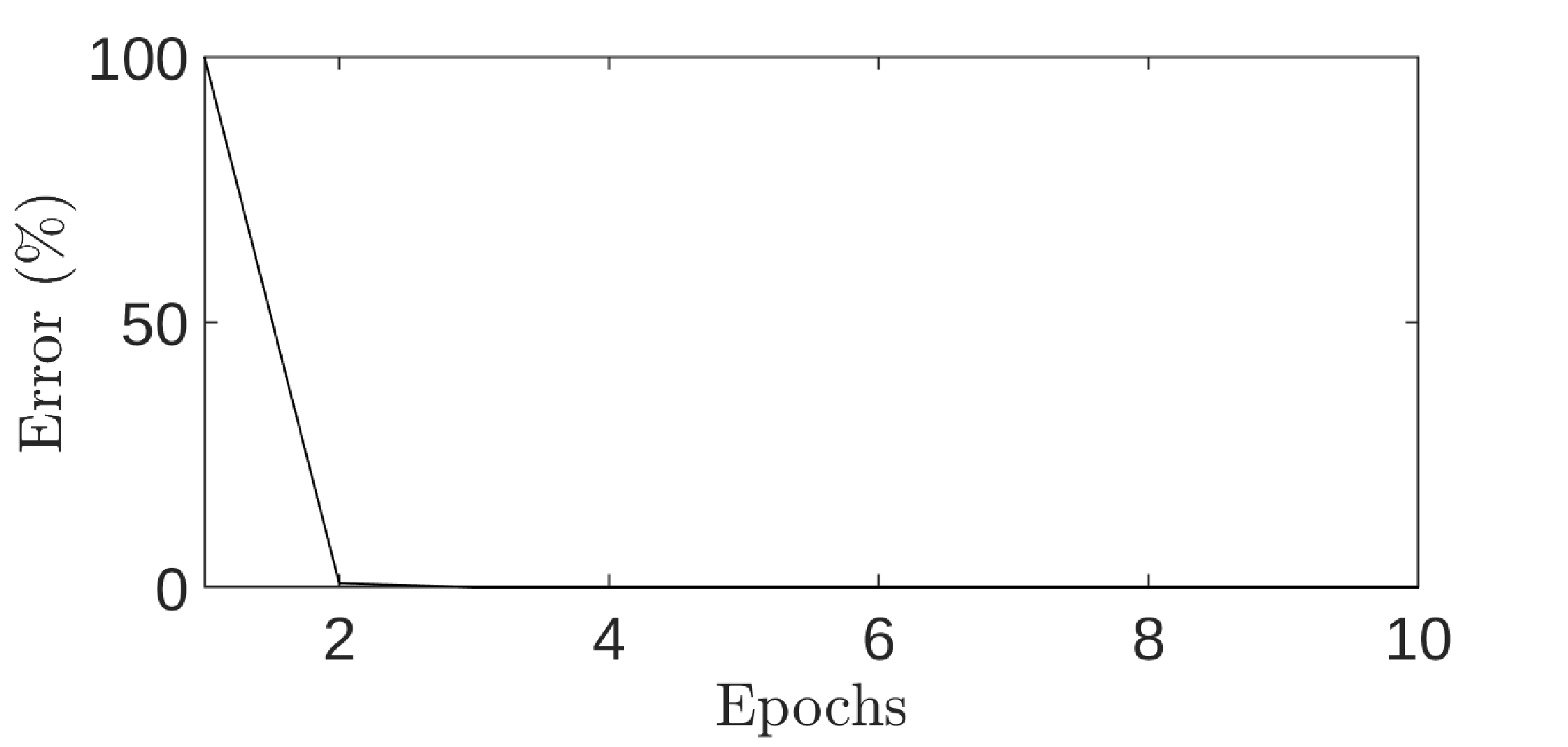}
    \includegraphics[width=0.495\textwidth]{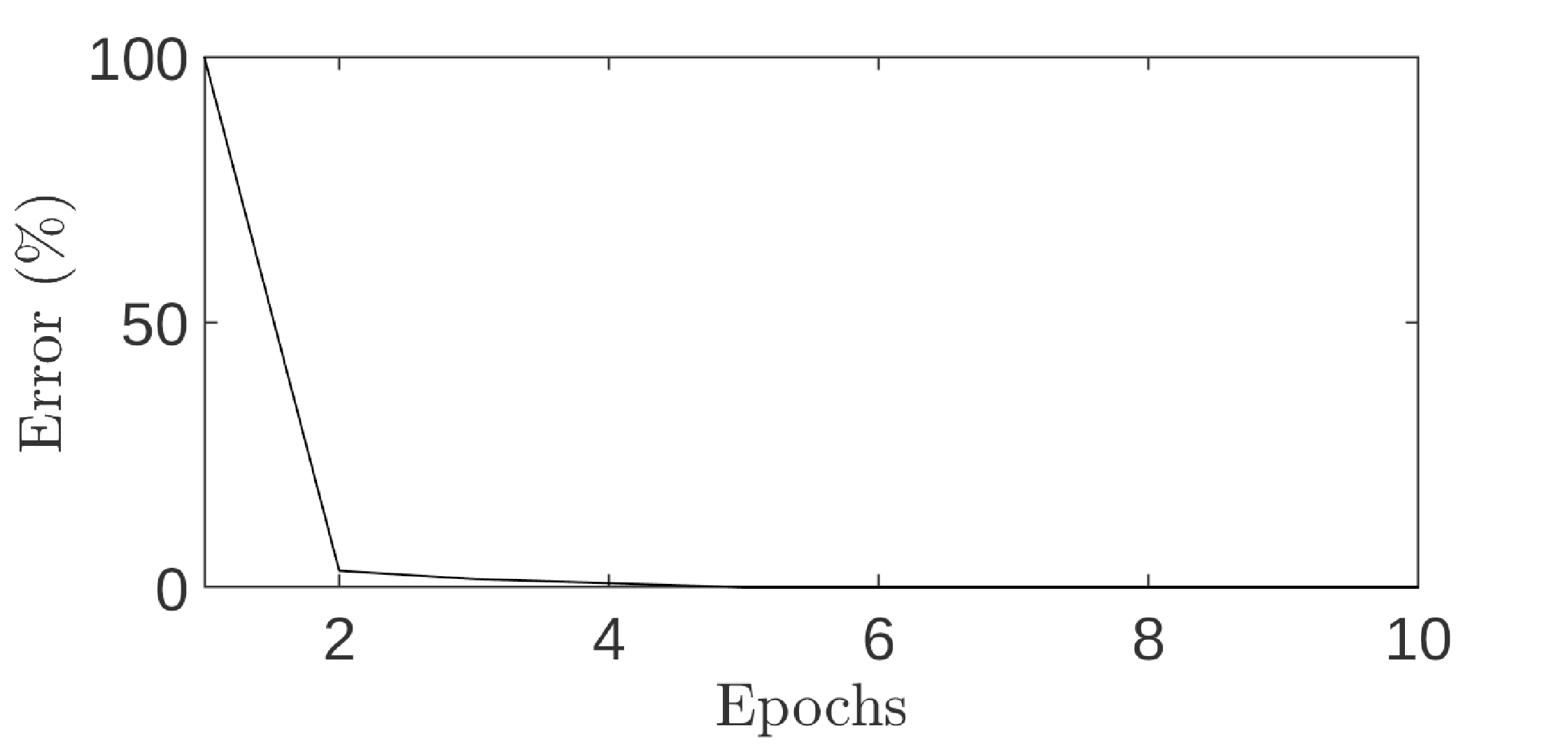}
\caption{Reconstruction error on the validation set over epochs of training for the encoder $E_1$ (left) and the encoder $E_2$ (right).}
\label{fig:res_sensors}       
\end{figure*}

Figure \ref{fig:res_sensors} shows the reconstruction error obtained by the encoders of the sensors on the validation set among epochs. It can be seen that around the fourth epoch for $E_x$, and sixth epoch for $E_y$, the reconstruction error converges to zero. At this point we consider that the encoders have learned to successfully represent any observation of the environment, and therefore, just one experiment was carried out for its training.

\begin{figure*}
    \centering
    \includegraphics[width=0.495\textwidth]{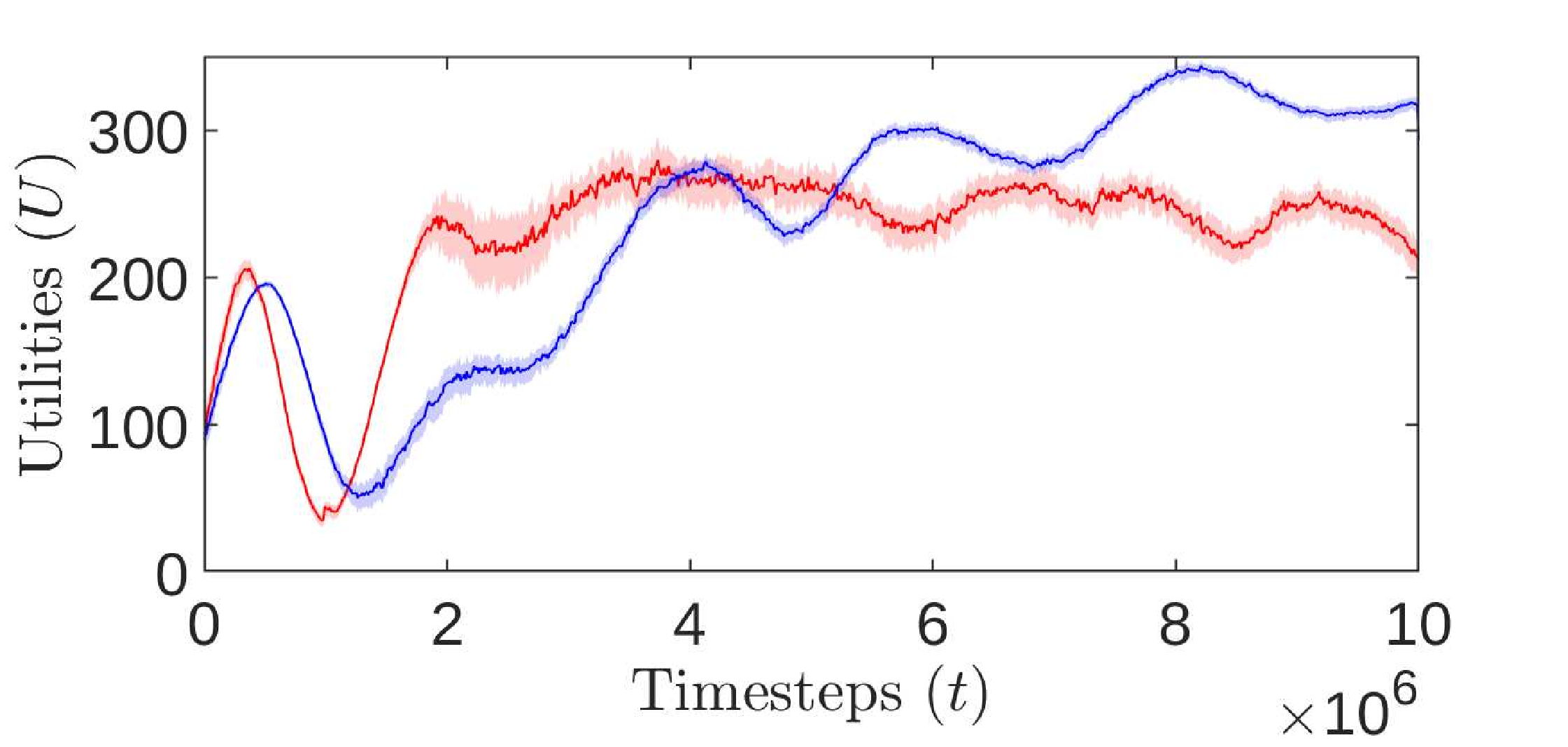}
    \includegraphics[width=0.495\textwidth]{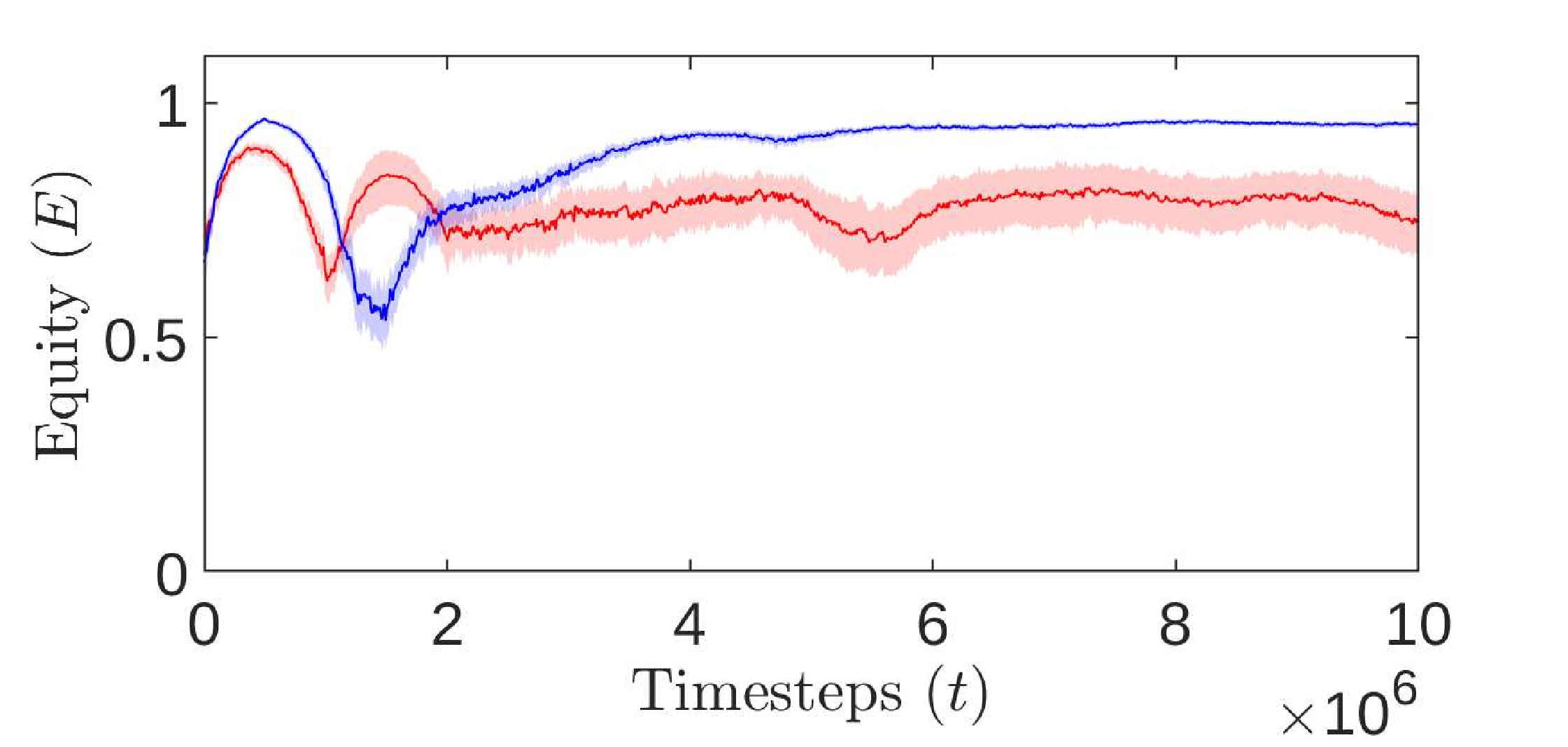}
    \includegraphics[width=0.495\textwidth]{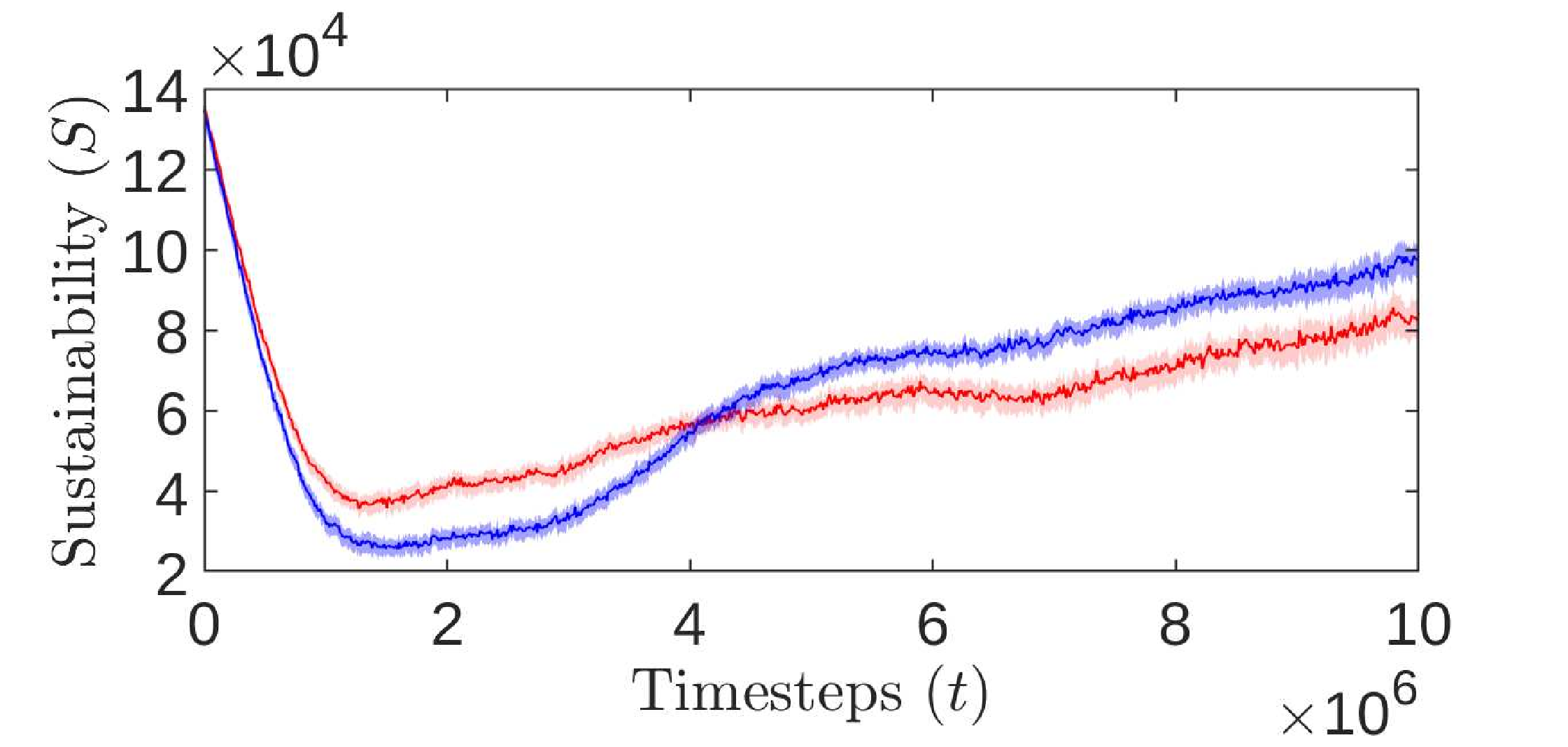}
    \includegraphics[width=0.495\textwidth]{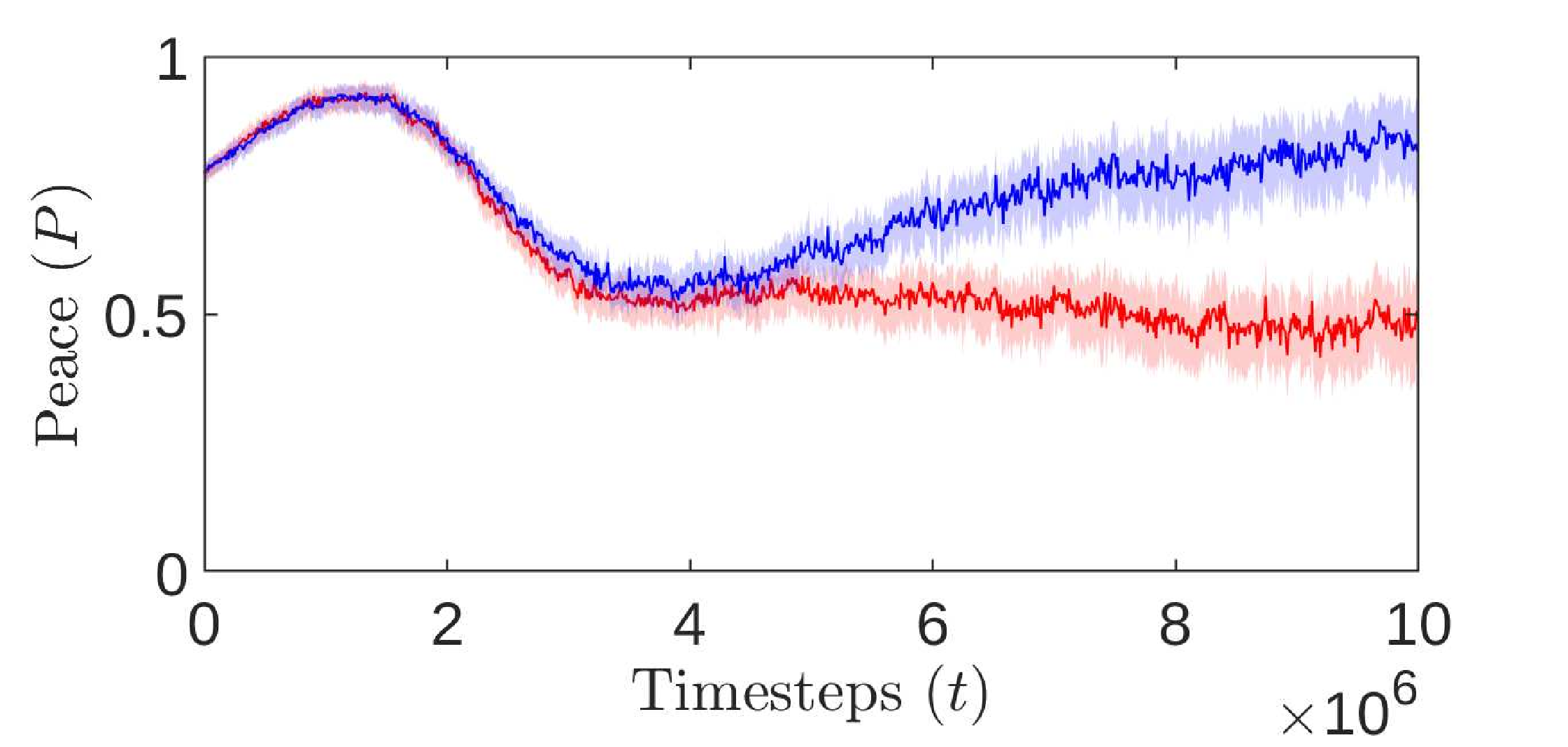}
    \includegraphics[width=0.495\textwidth]{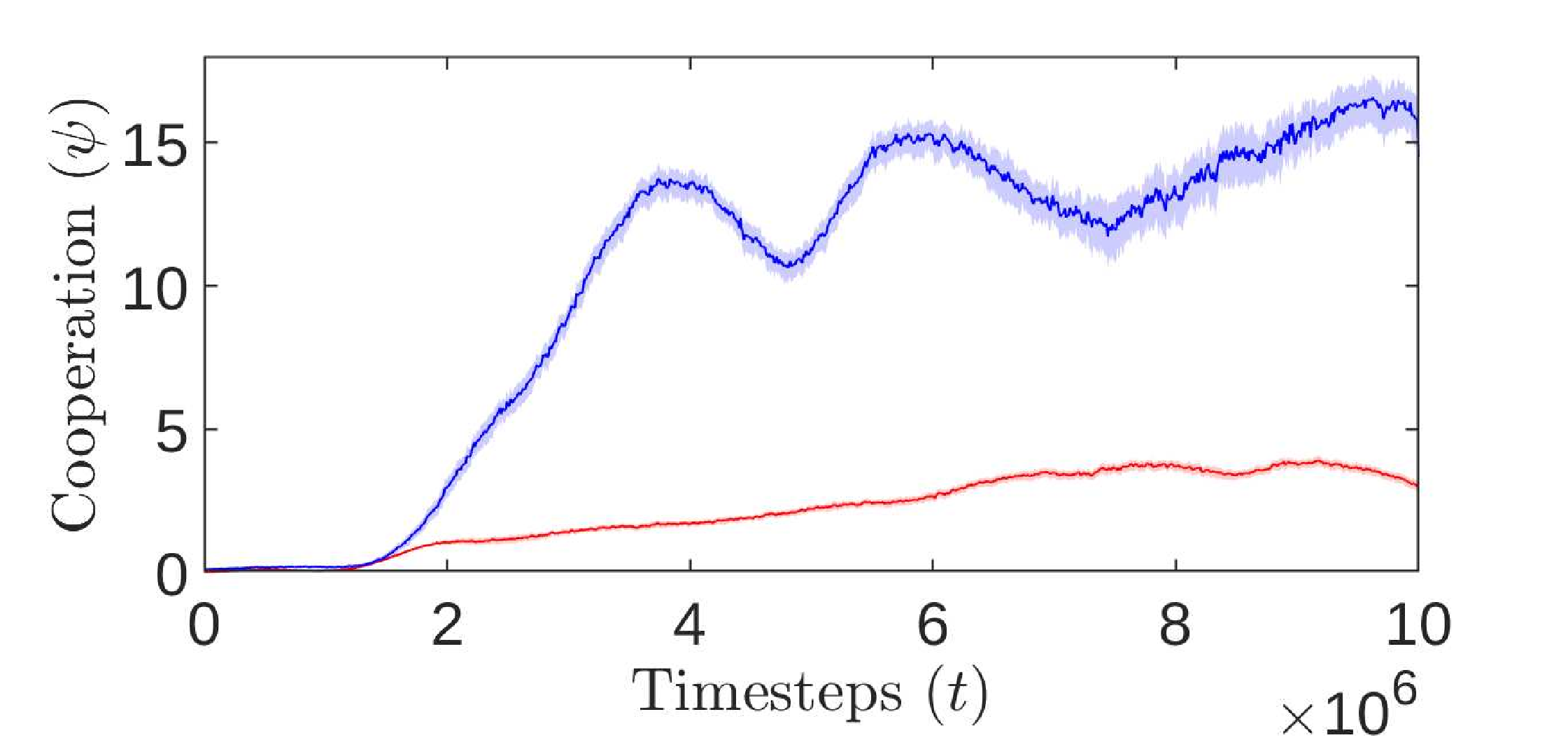}
\caption{Temporal evolution of the performance indices for the Correlation Maximizing System (blue), and the baseline system (red). The solid line is the sample mean obtained with 30 independent runs. The shaded area shows a confidence interval of 99.5 \%}
\label{fig:res}       
\end{figure*}

\begin{figure*}
    \centering
    \includegraphics[width=0.495\textwidth]{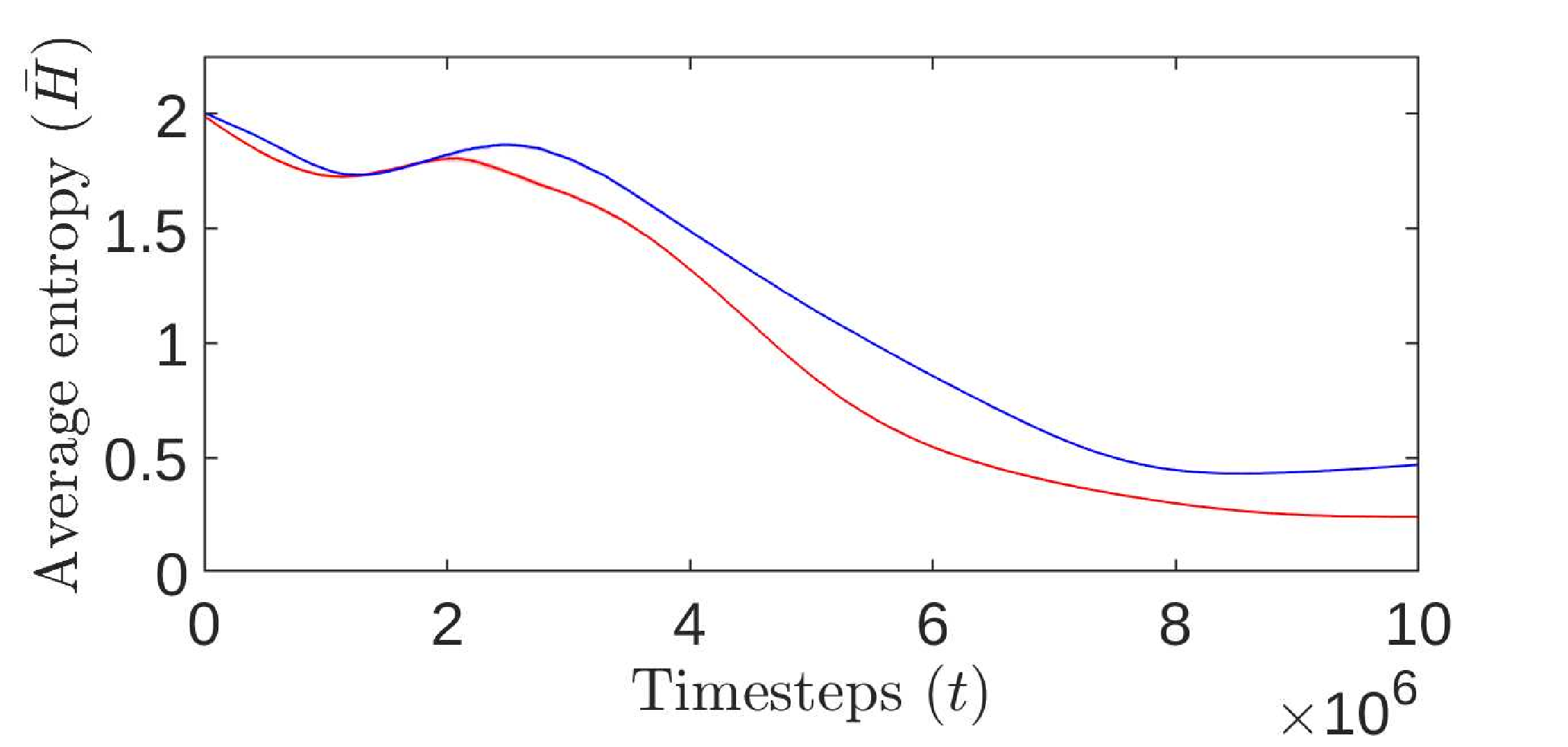}
    \includegraphics[width=0.495\textwidth]{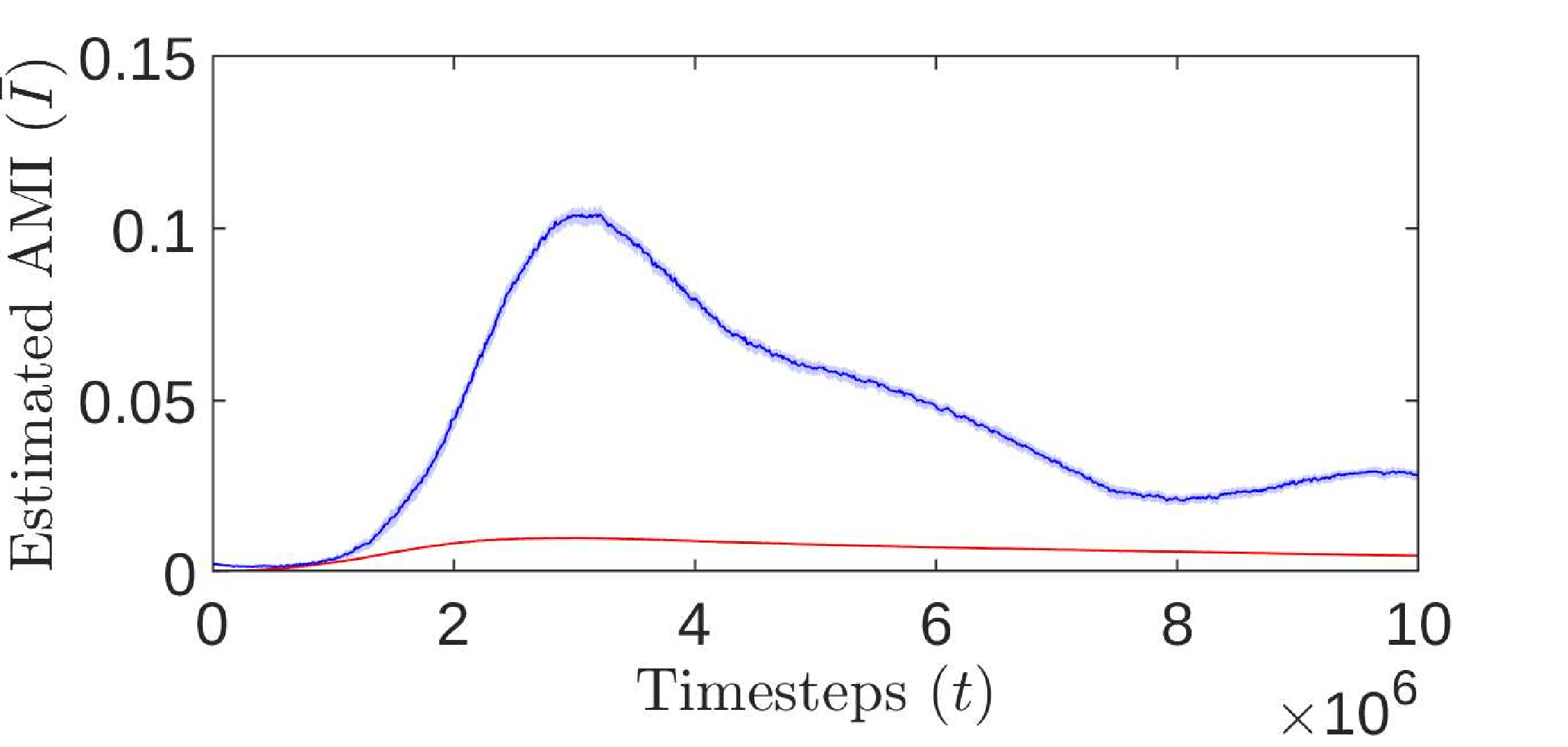}
\caption{Temporal evolution of the information for the Correlation Maximizing System (blue), and the baseline system (red). The solid line is the sample mean obtained with 30 independent runs. The shaded area shows a confidence interval of 99.5 \%}
\label{fig:res_info}       
\end{figure*}

Figure \ref{fig:res} depicts the temporal evolution of the performance indices across the 30 independent runs. These are calculated at the end of each iteration of the interaction loop in algorithm \ref{algo}. It can be seen that the CMS surpasses the baseline system on all the indices by the end of training. Table \ref{tab:res_sum} complements this results in terms of the initial and final values of the mean and variance of the performance indices. We also follow the dynamics of information as these can also provide valuable insight about the system. Figure \ref{fig:res_info} illustrates the temporal evolution of the average entropy and the EAMI across the 30 independent runs.


\begin{table}[!t]
\caption{Summary of results for the performance indices at the beginning and end of training. The best results at the end of training, in terms of higher mean and low variance, are shown in bold.}
\label{tab:res_sum}       
\centering
\begin{tabular}{@{}llllll@{}}
\hline\noalign{\smallskip}
&                    & \multicolumn{2}{l}{Baseline}   & \multicolumn{2}{l}{CMS}        \\ \hline\noalign{\smallskip}
&Index              & Mean                & Variance           &  Mean                & Variance           \\ \hline\noalign{\smallskip}
\multirow{5}{*}{\shortstack{Initial}} & Utilities       & 93.2728              & 358.9156              & 90.0706            & 321.3627             \\
&Equity        & 0.6778               & 0.0040               & 0.6618               & 0.0075               \\
&Peace            & 0.7732               & 0.0016               & 0.7749               & 0.0011              \\
&Sustainability & $1.3560 \times 10^5$ & $1.2317 \times 10^7$ & $1.3472 \times 10^5$ & $1.7866 \times 10^7$ \\
&Cooperation    & 0.0170               & 0.0001               & 0.0096              & $1.1727 \times 10^-5$             \\ \hline\noalign{\smallskip}

\multirow{5}{*}{\shortstack{Final}} & Utilities       & 211.8189              & 277.6377              & \textbf{293.4253}             & \textbf{52.8959}              \\
&Equity        & 0.7464               & 0.0125               & \textbf{0.9530}               & \textbf{0.0001}               \\
&Peace           & 0.5085               & \textbf{0.0266}               & \textbf{0.8157}               & 0.0284              \\
&Sustainability & $8.4526 \times 10^4$ & $6.0339 \times 10^7$ & $\mathbf{9.8540 \times 10^4}$ & $\mathbf{5.8366 \times 10^7}$ \\
&Cooperation    & 2.9994               & \textbf{0.0624}               & \textbf{14.5037}              & 1.9067             \\ \hline\noalign{\smallskip}
\end{tabular}
\end{table}

For the first $5 \times 10^5$ time steps the performance indices exhibit dynamics akin to the ones described in \cite{Perolat17a}. Initially, agents go through a learning phase during which they learn to harvest with increasing speed. This phase is characterized by the increase of the utility index and the descent of the sustainability index. The increase in the peace index indicates that agents learn not to use its beam, as the resource is still under-exploited, and there is no need to compete for it. The entropy of the policies decreases as these converge towards over-exploitative strategies. The EAMI, and hence the cooperation index, remain fairly low for both systems, since agents find no need to coordinate its actions, and the policy gradient outweighs the MI maximization term. 

At the end of this first phase, the speed with which agents harvest apples surpasses the regeneration rate of the resource and the utilities begin to decrease, reaching its minimum at $t=9.81\times 10^5$ and $t=1.23\times 10^6$, for the baseline and CMS, respectively. Passed this point, the environment turns competitive, and agents begin to use its beam against each other, resulting in the descent of the peace and equity indices, and the rising of the sustainability index. The average entropy and EAMI also begin to increase since the over-exploitative policies are no longer a good strategy and the agent faces this uncertain scenario in which should take into account the presence of other agents. The EAMI grows much faster for the CMS than for the baseline. Fitting a line to the EAMI in the interval between $t=1.5\times10^6$ and $t=2.5\times10^6$ results in a slope of $4.0258 \times 10^9$ for the baseline system and $7.0947 \times 10^8$ for the CMS. As a result of the increase in EAMI, the cooperation index also increases.

Both systems evolve similarly up to $t=2.72 \times 10^6$, although slower in the case of the CMS. From this point on, there are significant variations. While in the baseline system the peace index has a downward trend for the duration of training, ending with a mean value of $P=0.5085$, in the CMS rises, reaching a final mean value of $P=0.8157$. The equity index in the baseline system converges to a value close to the initial one, $E=0.7513$, whereas in the CMS increases up to a final mean value of $E=0.9554$, notably, with a much lower variance across experiments. In both systems the utilities, sustainability and cooperation indices have a growing trend, but for the CMS the final values, $U=293.4253$, $S=9.8541 \times 10^4$ and $\psi=14.507$, are higher than for the baseline system, $U=211.8189$, $S=8.4526 \times 10^4$ and $\psi=2.9994$. 


The information dynamics of both systems follows alike trends. The average entropy decreases and seems to be converging by the end of training. The EAMI initially increases, reaching its maximum value around $t = 3.073 \times 10^6$, and then decreases tending to convergence. The descend of EAMI follows the descend of entropy in the system since the minimum entropy of a set of random variables is an upper bound for its MI. Given that for the CMS the EAMI grows more than one order of magnitude faster, its maximum value, $\bar{I} = 0.1043$, is also much higher than in the baseline, $\bar{I} = 0.01$, which results in a big improvement in the cooperation index.

\section{Discussion}
\label{discussion}
The common initial dynamics of the CMS and baseline system, and its posterior divergence, could suggest that the search space of the CG has a region characterized by competitive low-correlated policies that, without the inclusion of the MI maximization term, is a local optimum surrounding the region of cooperative policies that is known to be the global optimum in SSDs \cite{Leibo17a}. This could be a characteristic of the optimization landscape of SSDs, which could explain the difficulties of traditional single-agent deep reinforcement learning algorithms to find optimal policies in such problems. The MI maximization term seems to modify the optimization landscape so that this region is no further a minimum and a gradient-based optimizer can find better solutions. The slower convergence of the CMS with respect to the baseline system could be explained considering that the maximization of MI also encourages the maximization of entropy, and hence, exploration.

The evidence provided in this work shows that the inclusion of a MI maximization term between the actions of the agents in its objective functions, results in a system with improved performance in the CG according to the utility, equity, peace and sustainability indices. The high values of these indices characterizes the behavior of a cooperative system in the CG \cite{Perolat17a}, therefore suggesting, in agreement with previous work \cite{Jaques18a}, and as captured by the proposed cooperation index, that high MI between agents is a characteristic of cooperative systems, and its maximization is a causal factor in the emergence of cooperation in the CG, and possibly, in general in social dilemmas. Extrapolating this idea to the many real world examples of social dilemmas that plague our society \cite{Capstick13a}\cite{Li14a}\cite{Johnson20a}, we could speculate that coordination and cooperation could emerge in such problems by implementing policies that promote the exchange of information between the parties involved.

\section{Conclusions}
\label{conclusions}
In this work we have proposed an index of cooperation in multi-agent systems as the product between the correlation of the actions of the agents and the global payoff of the system, and, based on this index, a deep reinforcement learning algorithm for the training of cooperative neural multi-agent systems. In addition to the estimation of the value and policy functions typically used to solve reinforcement learning problems, we also estimated the mutual information between the actions of the agents and, to promote coordination between them, introduced a term for its maximization in the learning problem. The proposed algorithm has the advantage of being decentralized, both in learning and execution, end-to-end differentiable, and scalable to populations of any size.

We applied the algorithm to the commons game, a problem that requires cooperation but in which traditional deep reinforcement learning algorithms struggle to find optimal solutions. The performance of our algorithm was compared according to multiple indices with the performance of a baseline system that does not maximize mutual information. The results showed that the system with maximization of mutual information consistently surpasses the baseline system on all indices. Based on this, we conclude that the maximization of mutual information between agents encourages the emergence of cooperation in the commons game.

The proposed algorithm makes several assumptions against which should be tested. Although, in principle, could deal with populations whose composition varies in time, this could affect convergence, as the $Y$ function would have to adapt to the changes to the joint policy product of the arrival and/or departure of agents. Our work also assumes that agents are homogeneous, since there is no way to tell them apart solely by observations. Verifying the robustness of the algorithm to variable populations and heterogeneity of agents is left as future work. One major limitation can also be the disentangling of information related to agents from information concerning the environment, required to train $E_y$. In the commons game this data is easily produced, but this will not hold for many other problems. Methods for unsupervised entity construction \cite{Garnelo16a} could help in this matter. Finally, is also worth to highlight that further experimentation should be done to test the robustness of the algorithm to the selection of its hyperparameters, such as the architecture of neural networks and parameters of the training algorithm.

\bibliographystyle{elsarticle-num}
\bibliography{references}

\begin{thebibliography}{10}
\expandafter\ifx\csname url\endcsname\relax
  \def\url#1{\texttt{#1}}\fi
\expandafter\ifx\csname urlprefix\endcsname\relax\def\urlprefix{URL }\fi
\expandafter\ifx\csname href\endcsname\relax
  \def\href#1#2{#2} \def\path#1{#1}\fi

\bibitem{AI100}
P.~Stone, R.~Brooks, E.~Brynjolfsson, R.~Calo, O.~Etzioni, G.~Hager,
  J.~Hirschberg, S.~Kalyanakrishnan, E.~Kamar, S.~Kraus, K.~Leyton-Brown,
  D.~Parkes, W.~Press, A.~Saxenian, J.~Shah, M.~Tambe, A.~Teller,
  \href{http://ai100.stanford.edu/2016-report}{Artificial intelligence and life
  in 2030}, Tech. rep., One Hundred Year Study on Artificial Intelligence:
  Report of the 2015-2016 Study Panel, Stanford University, Stanford, CA
  (September 2016).
\newline\urlprefix\url{http://ai100.stanford.edu/2016-report}

\bibitem{Schwarting19a}
W.~Schwarting, A.~Pierson, J.~Alonso-Mora, S.~Karaman, D.~Rus,
  \href{https://www.pnas.org/content/116/50/24972}{Social behavior for
  autonomous vehicles}, Proceedings of the National Academy of Sciences
  116~(50) (2019) 24972--24978 (2019).
\newblock \href
  {http://arxiv.org/abs/https://www.pnas.org/content/116/50/24972.full.pdf}
  {\path{arXiv:https://www.pnas.org/content/116/50/24972.full.pdf}}, \href
  {http://dx.doi.org/10.1073/pnas.1820676116}
  {\path{doi:10.1073/pnas.1820676116}}.
\newline\urlprefix\url{https://www.pnas.org/content/116/50/24972}

\bibitem{Panait05a}
L.~Panait, S.~Luke,
  \href{https://doi.org/10.1007/s10458-005-2631-2}{Cooperative multi-agent
  learning: The state of the art}, Autonomous Agents and Multi-Agent Systems
  11~(3) (2005) 387--434 (Nov. 2005).
\newblock \href {http://dx.doi.org/10.1007/s10458-005-2631-2}
  {\path{doi:10.1007/s10458-005-2631-2}}.
\newline\urlprefix\url{https://doi.org/10.1007/s10458-005-2631-2}

\bibitem{Sutton98a}
R.~S. Sutton, A.~G. Barto,
  \href{http://incompleteideas.net/book/the-book-2nd.html}{Reinforcement
  Learning: An Introduction}, 2nd Edition, The MIT Press, 2018.
\newline\urlprefix\url{http://incompleteideas.net/book/the-book-2nd.html}

\bibitem{Lecun15a}
Y.~LeCun, Y.~Bengio, G.~Hinton, \href{https://doi.org/10.1038/nature14539}{Deep
  learning}, Nature 521~(7553) (2015) 436--444 (2015).
\newblock \href {http://dx.doi.org/10.1038/nature14539}
  {\path{doi:10.1038/nature14539}}.
\newline\urlprefix\url{https://doi.org/10.1038/nature14539}

\bibitem{Hernandez19a}
P.~Hernandez-Leal, B.~Kartal, M.~E. Taylor,
  \href{https://doi.org/10.1007/s10458-019-09421-1}{A survey and critique of
  multiagent deep reinforcement learning}, Autonomous Agents and Multi-Agent
  Systems 33~(6) (2019) 750--797 (2019).
\newblock \href {http://dx.doi.org/10.1007/s10458-019-09421-1}
  {\path{doi:10.1007/s10458-019-09421-1}}.
\newline\urlprefix\url{https://doi.org/10.1007/s10458-019-09421-1}

\bibitem{Klyubin05a}
A.~S. {Klyubin}, D.~{Polani}, C.~L. {Nehaniv}, Empowerment: a universal
  agent-centric measure of control, in: 2005 IEEE Congress on Evolutionary
  Computation, Vol.~1, 2005, pp. 128--135 Vol.1.

\bibitem{Mohamed15a}
S.~{Mohamed}, D.~{Jimenez Rezende}, {Variational Information Maximisation for
  Intrinsically Motivated Reinforcement Learning}, arXiv e-prints (2015)
  arXiv:1509.08731 (Sep. 2015).
\newblock \href {http://arxiv.org/abs/1509.08731} {\path{arXiv:1509.08731}}.

\bibitem{Kumar18a}
N.~M. Kumar, \href{http://arxiv.org/abs/1810.05533}{Empowerment-driven
  exploration using mutual information estimation}, CoRR abs/1810.05533 (2018).
\newblock \href {http://arxiv.org/abs/1810.05533} {\path{arXiv:1810.05533}}.
\newline\urlprefix\url{http://arxiv.org/abs/1810.05533}

\bibitem{Jaques18a}
N.~Jaques, A.~Lazaridou, E.~Hughes, {\c{C}}.~G{\"{u}}l{\c{c}}ehre, P.~A.
  Ortega, D.~Strouse, J.~Z. Leibo, N.~de~Freitas,
  \href{http://arxiv.org/abs/1810.08647}{Intrinsic social motivation via causal
  influence in multi-agent {RL}}, CoRR abs/1810.08647 (2018).
\newblock \href {http://arxiv.org/abs/1810.08647} {\path{arXiv:1810.08647}}.
\newline\urlprefix\url{http://arxiv.org/abs/1810.08647}

\bibitem{Lindenfors17a}
P.~Lindenfors, \href{https://doi.org/10.1007/978-3-319-50874-0}{For Whose
  Benefit?}, Springer International Publishing, 2017.
\newblock \href {http://dx.doi.org/10.1007/978-3-319-50874-0}
  {\path{doi:10.1007/978-3-319-50874-0}}.
\newline\urlprefix\url{https://doi.org/10.1007/978-3-319-50874-0}

\bibitem{Murphey02a}
R.~Murphey, \href{https://doi.org/10.1007/0-306-47536-7_9}{An Introduction to
  Collective and Cooperative Systems}, Springer US, Boston, MA, 2002, pp.
  171--197.
\newblock \href {http://dx.doi.org/10.1007/0-306-47536-7_9}
  {\path{doi:10.1007/0-306-47536-7_9}}.
\newline\urlprefix\url{https://doi.org/10.1007/0-306-47536-7_9}

\bibitem{Griffith14a}
V.~Griffith, C.~Koch,
  \href{https://doi.org/10.1007/978-3-642-53734-9_6}{Quantifying Synergistic
  Mutual Information}, Springer Berlin Heidelberg, Berlin, Heidelberg, 2014,
  pp. 159--190.
\newblock \href {http://dx.doi.org/10.1007/978-3-642-53734-9_6}
  {\path{doi:10.1007/978-3-642-53734-9_6}}.
\newline\urlprefix\url{https://doi.org/10.1007/978-3-642-53734-9_6}

\bibitem{Ha18a}
D.~Ha, J.~Schmidhuber, \href{http://arxiv.org/abs/1803.10122}{World models},
  CoRR abs/1803.10122 (2018).
\newblock \href {http://arxiv.org/abs/1803.10122} {\path{arXiv:1803.10122}}.
\newline\urlprefix\url{http://arxiv.org/abs/1803.10122}

\bibitem{Konda03a}
V.~R. Konda, J.~N. Tsitsiklis, Onactor-critic algorithms, SIAM J. Control and
  Optimization 42, pp. 1143-1166 (2003).

\bibitem{Goodfellow16a}
I.~Goodfellow, Y.~Bengio, A.~Courville, Deep Learning, The MIT Press, 2016.

\bibitem{Aberdeen03a(revised)}
D.~Aberdeen, A (revised) survey of approximate methods for solving partially
  observable markov decision processes, Tech. rep., rep., National ICT
  Australia (2003).

\bibitem{Jaeger01a}
H.~Jaeger,
  \href{http://www.faculty.jacobs-university.de/hjaeger/pubs/EchoStatesTechRep.pdf}{The
  "echo state" approach to analysing and training recurrent neural networks},
  GMD Report 148, GMD - German National Research Institute for Computer Science
  (2001).
\newline\urlprefix\url{http://www.faculty.jacobs-university.de/hjaeger/pubs/EchoStatesTechRep.pdf}

\bibitem{hjelm18a}
R.~{Devon Hjelm}, A.~{Fedorov}, S.~{Lavoie-Marchildon}, K.~{Grewal},
  P.~{Bachman}, A.~{Trischler}, Y.~{Bengio}, {Learning deep representations by
  mutual information estimation and maximization}, arXiv e-prints (2018)
  arXiv:1808.06670 (Aug 2018).
\newblock \href {http://arxiv.org/abs/1808.06670} {\path{arXiv:1808.06670}}.

\bibitem{Schulman17a}
J.~Schulman, F.~Wolski, P.~Dhariwal, A.~Radford, O.~Klimov,
  \href{http://arxiv.org/abs/1707.06347}{Proximal policy optimization
  algorithms}, CoRR abs/1707.06347 (2017).
\newblock \href {http://arxiv.org/abs/1707.06347} {\path{arXiv:1707.06347}}.
\newline\urlprefix\url{http://arxiv.org/abs/1707.06347}

\bibitem{Leibo17a}
J.~Z. Leibo, V.~Zambaldi, M.~Lanctot, J.~Marecki, T.~Graepel, Multi-agent
  reinforcement learning in sequential social dilemmas, in: Proceedings of the
  16th Conference on Autonomous Agents and MultiAgent Systems, AAMAS ’17,
  International Foundation for Autonomous Agents and Multiagent Systems,
  Richland, SC, 2017, p. 464–473.

\bibitem{Perolat17a}
J.~P{\'{e}}rolat, J.~Z. Leibo, V.~F. Zambaldi, C.~Beattie, K.~Tuyls,
  T.~Graepel, \href{http://arxiv.org/abs/1707.06600}{A multi-agent
  reinforcement learning model of common-pool resource appropriation}, CoRR
  abs/1707.06600 (2017).
\newblock \href {http://arxiv.org/abs/1707.06600} {\path{arXiv:1707.06600}}.
\newline\urlprefix\url{http://arxiv.org/abs/1707.06600}

\bibitem{Kingma14a}
D.~P. Kingma, J.~Ba, \href{http://arxiv.org/abs/1412.6980}{Adam: A method for
  stochastic optimization}, cite arxiv:1412.6980Comment: Published as a
  conference paper at the 3rd International Conference for Learning
  Representations, San Diego, 2015 (2014).
\newline\urlprefix\url{http://arxiv.org/abs/1412.6980}

\bibitem{Capstick13a}
S.~B. Capstick, Public understanding of climate change as a social dilemma,
  Sustainability 5~(8) (2013) 3484--3501 (2013).

\bibitem{Li14a}
Y.~Li, F.~K. Yao, D.~Ahlstrom,
  \href{https://doi.org/10.1007/s10490-014-9406-8}{The social dilemma of
  bribery in emerging economies: A dynamic model of emotion, social value, and
  institutional uncertainty}, Asia Pacific Journal of Management 32~(2) (2014)
  311--334 (Dec. 2014).
\newblock \href {http://dx.doi.org/10.1007/s10490-014-9406-8}
  {\path{doi:10.1007/s10490-014-9406-8}}.
\newline\urlprefix\url{https://doi.org/10.1007/s10490-014-9406-8}

\bibitem{Johnson20a}
T.~Johnson, C.~Dawes, J.~Fowler, O.~Smirnov,
  \href{https://doi.org/10.30636/jbpa.31.150}{Slowing {COVID}-19 transmission
  as a social dilemma: Lessons for government officials from interdisciplinary
  research on cooperation}, Journal of Behavioral Public Administration 3~(1)
  (Apr. 2020).
\newblock \href {http://dx.doi.org/10.30636/jbpa.31.150}
  {\path{doi:10.30636/jbpa.31.150}}.
\newline\urlprefix\url{https://doi.org/10.30636/jbpa.31.150}

\bibitem{Garnelo16a}
M.~Garnelo, K.~Arulkumaran, M.~Shanahan,
  \href{http://arxiv.org/abs/1609.05518}{Towards deep symbolic reinforcement
  learning}, CoRR abs/1609.05518 (2016).
\newblock \href {http://arxiv.org/abs/1609.05518} {\path{arXiv:1609.05518}}.
\newline\urlprefix\url{http://arxiv.org/abs/1609.05518}

\end{thebibliography}

%
%
\end{document}